\documentclass[conference]{IEEEtran}
\IEEEoverridecommandlockouts
\usepackage{cite}
\usepackage{multirow}
\usepackage{amsmath,amssymb,amsfonts}
\usepackage{algorithmic}
\usepackage{graphicx}
\usepackage{textcomp}
\usepackage{xcolor}
\usepackage{booktabs}
\def\BibTeX{{\rm B\kern-.05em{\sc i\kern-.025em b}\kern-.08em
    T\kern-.1667em\lower.7ex\hbox{E}\kern-.125emX}}

\begin{document}

\title{Spatio-Temporal Data Mining for\\ Aviation Delay Prediction 
}

\author{\IEEEauthorblockN{Kai Zhang\textsuperscript{*} , Yushan Jiang\textsuperscript{*}, Dahai Liu\textsuperscript{†}, Houbing Song\textsuperscript{*}}
\IEEEauthorblockA{\textsuperscript{*}Department of Electrical Engineering and Computer Science, Embry-Riddle Aeronautical University \\
\textsuperscript{†}College of Aviation, Embry-Riddle Aeronautical University\\
Email: zhangk3@my.erau.edu, jiangy2@my.erau.edu, dahai.liu@erau.edu, houbing.song@erau.edu}
}

\maketitle
\begin{abstract}
To accommodate the unprecedented increase of commercial airlines over the next ten years, the Next Generation Air Transportation System (NextGen) has been implemented in the USA that records large-scale Air Traffic Management (ATM) data to make air travel safer, more efficient, and more economical. A key role of collaborative decision making for air traffic scheduling and airspace resource management is the accurate prediction of flight delay. There has been a lot of attempts to apply data-driven methods such as machine learning to forecast flight delay situation using air traffic data of departures and arrivals. However, most of them omit en-route spatial information of airlines and temporal correlation between serial flights which results in inaccuracy prediction. In this paper, we present a novel aviation delay prediction system based on stacked Long Short-Term Memory (LSTM) networks for commercial flights. The system learns from historical trajectories from automatic dependent surveillance-broadcast (ADS-B) messages and uses the correlative geolocations to collect indispensable features such as climatic elements, air traffic, airspace, and human factors data along posterior routes. These features are integrated and then are fed into our proposed regression model. The latent spatio-temporal patterns of data are abstracted and learned in the LSTM architecture. Compared with previous schemes, our approach is demonstrated to be more robust and accurate for large hub airports. 

\end{abstract}

\begin{IEEEkeywords}
Data mining, multivariate statistics, LSTM, ADS-B, ATM.
\end{IEEEkeywords}

\section{Introduction}
Accurate prediction of aviation delay for commercial flights is a key component to improve the safety, capacity, efficiency in air traffic management, and airline business \cite{song2017smart,song2017security,7406686,song2016cyber,7926913}. However, as a dynamic system with uncertainties, civil aviation faces multiple unexpected events frequently that can result in flight delays or cancellations. In the case of the flight delay propagation between initially delayed flight and flights downstream, the impact can grow exponentially if the air traffic control cannot adopt an appropriate resource re-allocation strategy to optimize mobility in time. Hence, estimating delay as accurately as possible within a controllable time window is critical. 

Traditional methods of the aviation delay prediction are relying on modeling and simulation techniques, but it is hard to select suitable modeling assumptions for the quality of the analysis \cite{kim2016deep}. With the large-scale deployments and applications of Internet of Things (IoT) devices, fast collection of ubiquitous information including spatial and temporal data motivates the utility of big data analytics and statistical machine learning in many fields\cite{lv2017next, dartmann2019big, liang2018deep, cai2019trading, zheng2020privacy, 8445615, tan2019low, 7545816}. A few studies on aviation delay prediction based on supervised machine learning have been conducted in recent years. For instance, Manna et al. \cite{manna2017statistical} employ Gradient Boosted Decision Tree to predict average arrival and departure delays of a day. However, the outliers in the dataset are removed directly before the training process so that the data completeness and model generalization is not guaranteed. Gopalakrishnan et al. \cite{gopalakrishnan2017comparative} compare the performance of Markov Jump Linear System, Classification and Regression Trees, and Artificial Neural Networks to predict delays in air traffic networks, and reveal a trade-off between model simplicity and prediction accuracy. Although promising results are obtained by using statistical machine learning methods, the crucial temporal elements, in particular delay propagation effect, cannot be learned by such one-shot prediction models.

Inspired by the considerable success of Recurrent Neural Networks (RNNs) and its variants applied in sequential event prediction such as natural language processing \cite{yin2017comparative}, few researchers implement RNNs to capture temporal correlation of factors which may potentially influence aviation for more accurate delay prediction \cite{kim2016deep, gui2019flight, mccarthy2019amsterdam, ayhan2018predicting, 10.1002/ett.3482, KONG2019460}. However, a lot of valuable attributes in spatial and temporal domains are not or rarely taken into account in these studies. Therefore, we present a spatio-temporal data mining framework based on stacked LSTM networks for aviation delay prediction to bridge this gap. By combining spatio-temporal features from available data sets containing flight path information, airspace characteristics and weather, our model can learn representations of spatio-temporal sequences with multiple levels of abstraction to predict subsequent aviation delay of an airport. To alleviate the overfitting problem, a regularization technique called Dropout \cite{zaremba2014recurrent} is applied in the middle of two adjacent stacked LSTM layers. 

Compared with previous work based on RNN, this paper makes three main contributions:
\begin{itemize}
    \item Unlike other systems that predict delay before departure \cite{ayhan2018predicting} or make an implicit assumption that the journey of each aircraft does not vary significantly \cite{mccarthy2019amsterdam}, we fully consider the indeterminacy of air transportation system to figure out the aviation delay prediction problem pointedly over a time horizon of one hour.
    \item We present a complete workflow of data manipulation in this paper. Some work \cite{ayhan2018predicting, gui2019flight} provide comparative analysis for delay prediction using several models including RNNs but they focus more on non-time-series methods especially boosting and bagging methods, whereas the input format for RNNs is a far away from the input format of other algorithms. Moreover, though \cite{kim2016deep} describes a data processing pipeline in detail, it creates day-to-day sequences rather than flight-to-flight or minute-to-minute sequences like us, which does not make sense in real-time air traffic control.
    \item Multi-source data are integrated and extended to generate a spatio-temporal dataset with richer information in spatial and temporal domains.  
\end{itemize}

The remainder of this paper is structured as follows. Section \uppercase\expandafter{\romannumeral2} formulates the aviation delay prediction problem and reviews the fundamental elements of the LSTM network. Section \uppercase\expandafter{\romannumeral3} introduces the data sets we use and the corresponding feature engineering. Then, our proposed LSTM-based architecture is introduced. The experimental setup and results are demonstrated in Section \uppercase\expandafter{\romannumeral4} that show the accuracy of our solution, while Section \uppercase\expandafter{\romannumeral5} concludes this paper and contemplates some future work.

\section{Problem Statement and LSTM Network}
Aviation delay is affected by multiple factors. Some of them are unpredictable such as military training activities, equipment failures at the airport, extreme weather. However, there are still valuable temporal contexts that can be retrieved for delay prediction. For example, the late arrival or departure of a previous flight will affect the on-time departure and arrival of succeeding flights. Such a pattern motivates us to model the aviation delay as a multivariate time series problem.

\subsection{Problem Formulation}
We select an airport $\mathcal{A}$ and flights $\mathcal{F}$ flying to $\mathcal{A}$ to formulate the arrival delay estimation at an airport. 

\begin{figure}
    \centering
    \includegraphics [scale =.25] {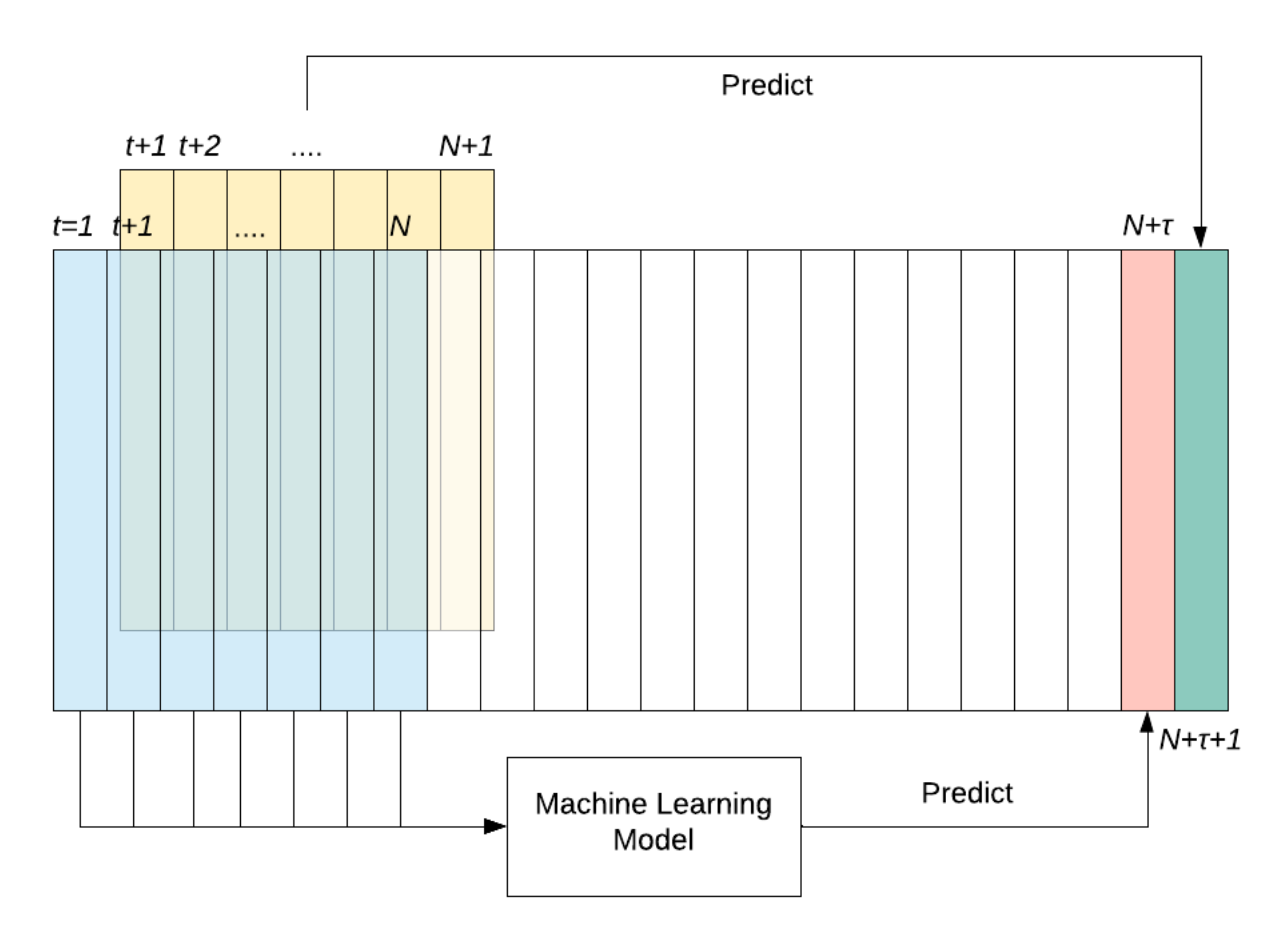}
    \caption{Transform time series data into a supervised learning problem.}
    \label{fig:problem}
\end{figure}

Considering the prediction function $f(X_{\mathcal{F}}^{\mathcal{A}}) \rightarrow Y_{\mathcal{F}}^{\mathcal{A}}$, with the input $X_{\mathcal{F}}^{\mathcal{A}}=\left\{x_{t}\right\}_{t=1}^{N}$ for $x \in \mathcal{R}^{M}$, where  (i) $N$ denotes the number of time stamps, (ii) $M$ denotes the number of variables or feature dimension, and (iii) $Y_{\mathcal{F}}^{\mathcal{A}}$ is the ground truth which is denoted as $Y_{\mathcal{F}}^{\mathcal{A}} = \left\{y_{T}\right\}_{T=t+N-1+\tau}$ where $y \in \mathcal{R}$, and $\tau$ is the self-defined time interval. Therefore,  our multivariate sequential data are transformed into a supervised learning problem. To be specific, if we use a sequence starting at $t=1$, the information in a certain period ($t \in [1,N]$ and $t \in \mathcal{Z}$) would be used to predict the delay at time stamp $N+\tau$ as shown in Fig. \ref{fig:problem}.

\subsection{LSTM}
Vanilla RNNs are elaborate for sequential prediction problem but it has difficulty in overcoming exploding and vanishing error flow to learn long-term dependence \cite{hochreiter1997long}. Therefore, a variant of RNNs named LSTM is well designed to address these problems and has made significant advancements in applications \cite{8685670,dartmann2019big}. In this paper, we feed our data to the LSTM model whose basic structure of a cell is shown in Fig. \ref{fig:lstmcell}. The gate mechanism is a key component in the LSTM structure. Gates are a track to optionally let information (hidden state $h_{t}$ and cell state $C_{t}$) through. There are three gates in an LSTM cell, Forget Gate, Input Gate and Output Gate, and their functions are represented by the following equations:

\begin{figure}
    \centering
    \includegraphics[width=0.45\textwidth] {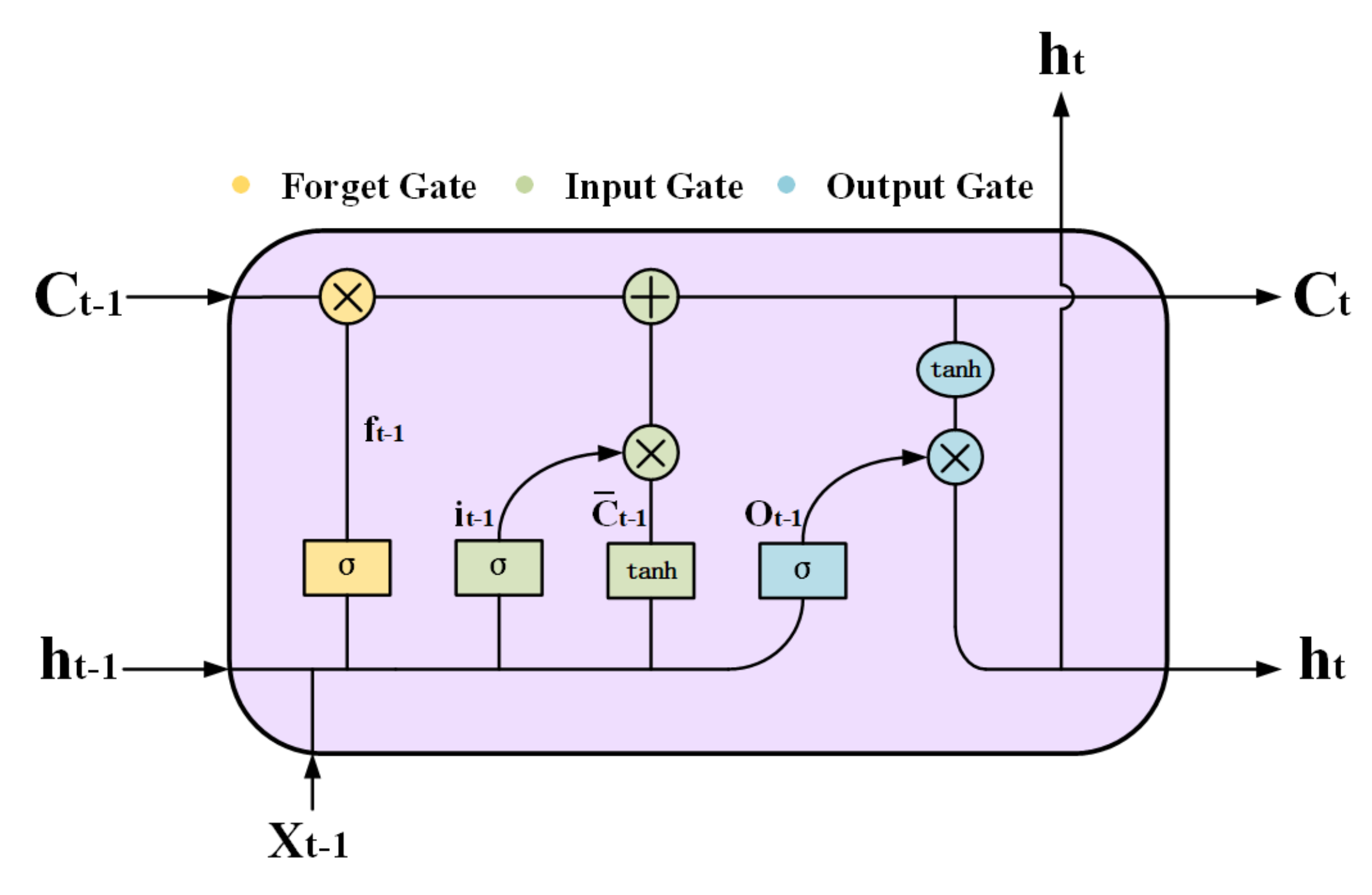}
    \caption{A LSTM cell. }
    \label{fig:lstmcell}
\end{figure}

\begin{figure*}
    \centering
    \includegraphics [scale =.5] {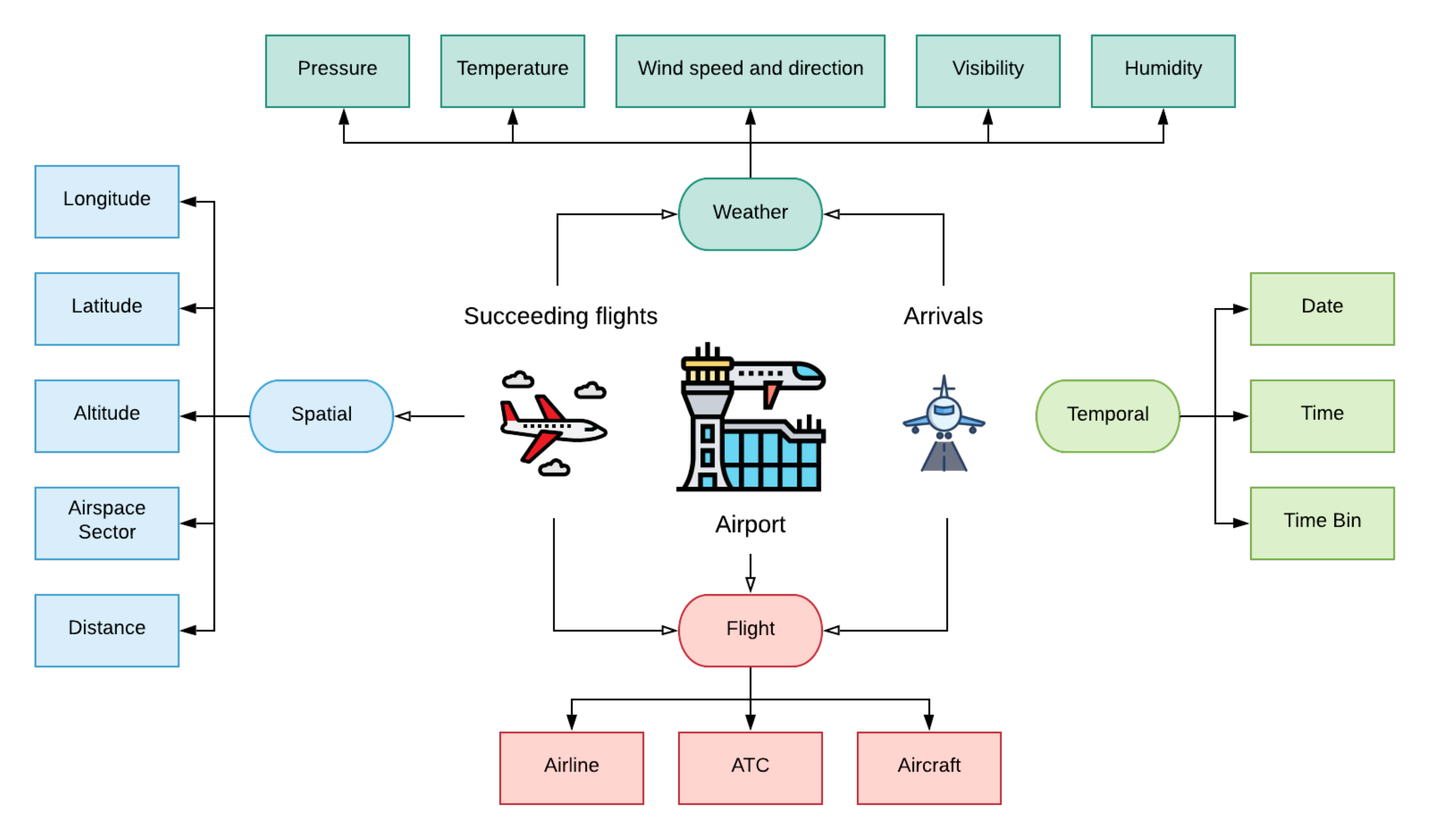}
    \caption{The overview of features. Once a flight arrives, it triggers the prediction mechanism based on airport information and the corresponding succeeding flight information. Various features in each domain can be combined concerning spatio-temporal characteristics. For example, combining weather and spatial features of succeeding flights produces a series of indicators that reflect the weather conditions of each aircraft's geographic locations. Then, if we consider temporal elements further, the synthetic features describe variations of flights over space and time synchronously.}
    \label{fig:features}
\end{figure*}

\begin{equation}
    f_{t}=\sigma\left(W_{f} \cdot\left[h_{t-1}, x_{t}\right]+b_{f}\right)
\end{equation}
\begin{equation}
i_{t} =\sigma\left(W_{i} \cdot\left[h_{t-1}, x_{t}\right]+b_{i}\right)
\end{equation}
\begin{equation}
\tilde{C}_{t} =\tanh \left(W_{C} \cdot\left[h_{t-1}, x_{t}\right]+b_{C}\right)
\end{equation}
\begin{equation}
    C_{t}=f_{t} * C_{t-1} + i_{t} * \tilde{C}_{t}
    \label{eq:cellstate}
\end{equation}
\begin{equation}
    o_{t}=\sigma\left(W_{o}\left[h_{t-1}, x_{t}\right]+b_{o}\right)
\end{equation}
\begin{equation}
    h_{t}=o_{t} * \tanh \left(C_{t}\right)
\end{equation}

\noindent where $\sigma$ is the sigmoid activation function which observes states at the previous step and outputs a number $f_{t}$ between 0 and 1 to control the degree of remaining information flow. $i_{t}$ is the input gate that is combined with the candidate value, $\tilde{C}_{t}$ after tanh layer to update the state, then the old cell state $C_{t}$ representing long-term memory can be replaced by the new one as shown in Equation \ref{eq:cellstate}. $h_{t}$ is the final output calculated by the multiplication of $o_{t}$ and tanh($C_{t}$) which could be the input for next LSTM cell.

\section{Methodology}
In this section, data sets used for aviation delay prediction are introduced, followed by feature engineering that constructs a proper set representing the underlying problem of the predictive task for improved model generalization. Besides, the design of our proposed LSTM-based architecture is explained.

\subsection{Data Description}
In this paper, we use data of the first day of each month for the period of July 2016 through February 2017 in Coordinated Universal Time (UTC). Note that, to guarantee the spatio-temporal continuity of data, only records whose \textit{Arrival Airport} is Hartsfield-Jackson Atlanta International Airport (ATL) are selected. The outline descriptions of features extracted from each domain are illustrated in Fig. \ref{fig:features}.

\begin{figure}
    \centering
    \includegraphics [width=0.35\textwidth] {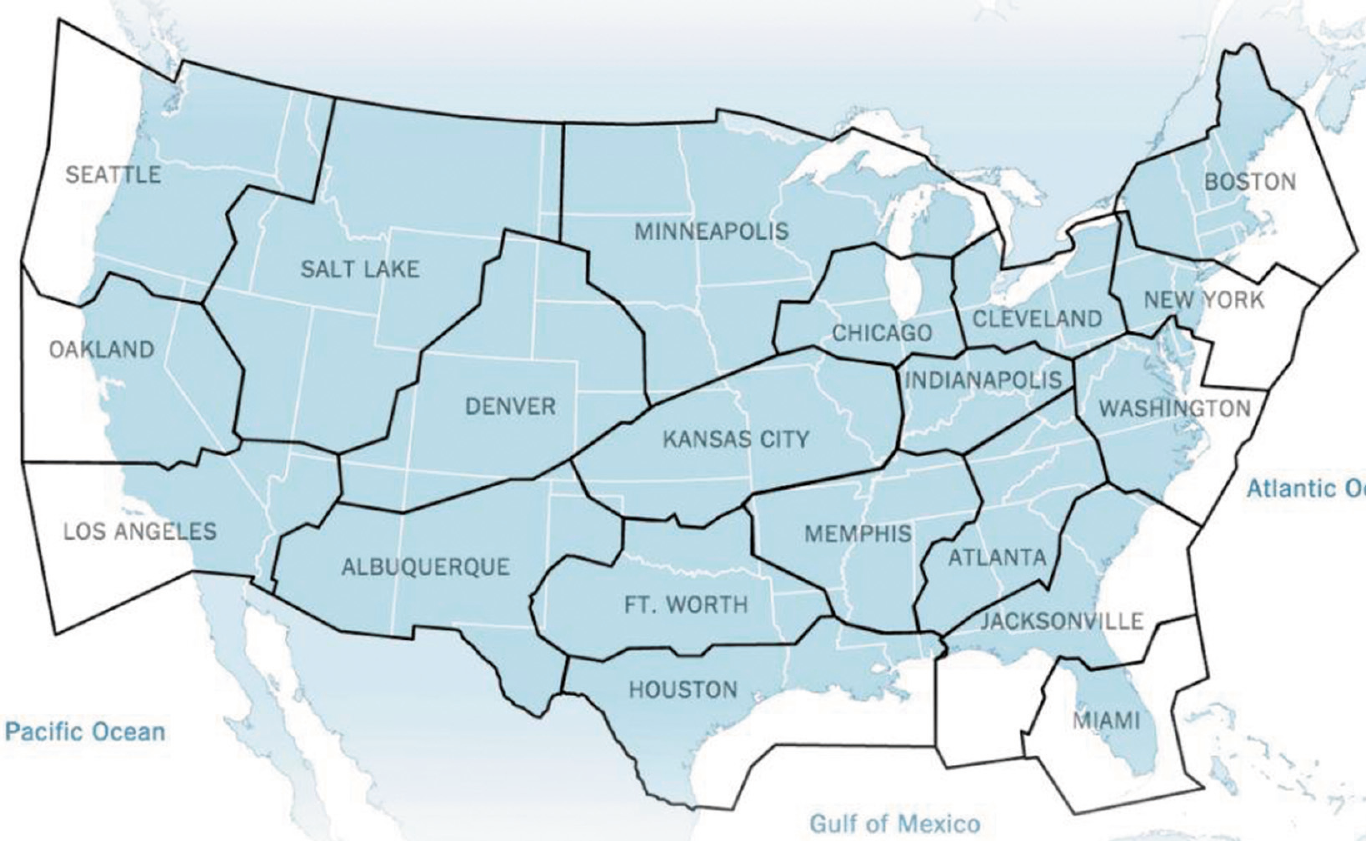}
    \caption{The boundaries of the 22 ARTCC system \cite{avweb}.}
    \label{fig:artcc}
\end{figure}

\subsubsection{Flight data} The flight records including delay information are provided by United States Department of Transportation. The data contains some most important information, for example, \textit{Departure/Arrival Airport}, \textit{Scheduled Departure/Arrival Time}, \textit{Actual Departure/Arrival Time} and \textit{Airlines}. 

\subsubsection{Trajectory data} In the United States, the aircraft's trajectories are collected continuously by automatic dependent surveillance-broadcast (ADS-B) \cite{scovel2013faa}, a surveillance technology used by Federal Aviation Administration (FAA) for air traffic control (ATC). The data from ADS-B Exchange \cite{exchange2018ads} contains all flights installed with ADS-B equipment which covers most commercial aviation. The fields of the ADS-B data includes \textit{Aircraft Identifications}, position information (\textit{Longitude, Latitude, Altitude}), flight status (such as \textit{Aircraft Speed} and \textit{Track Angle}) and aircraft attributes (such as \textit{Aircraft Model} and \textit{Manufacturer's Name}).

\subsubsection{ATC data}
En-route flights are governed by air route traffic control centers (ARTCCs) whose division is depicted in Fig \ref{fig:artcc},  which is the result of historical traffic growth and population. The information of each ARTCC facility such as \textit{Staffing} and \textit{Controller Training Time}, can be obtained from 123ATC \cite{123atc}.

\subsubsection{Weather data}
The weather condition in the air route is demonstrated as a significant factor for the delay prediction task. Therefore, we gather Local Climatological Data (LCD) from the National Oceanic and Atmospheric Administration (NOAA). The flight-related data involves \textit{Temperature, Precipitation, Humidity, Sky Conditions, Wind Speed, Wind Direction} and so on. 

\subsection{Feature Engineering}
As a preprocessing step, feature engineering refers to the process of using knowledge and statistical approaches to select and create representative features. We aim to construct a cleansed data set with relatively high-dimensional feature space after feature engineering, so it enables the trained model to have better performance on unseen data.

\subsubsection{Feature selection} 
Features with the percentage of missing values that exceed the threshold (80\%) are removed. Then, we calculate the Pearson correlations of each pairwise variable and remove redundant features beyond the threshold (80\%). Among the remaining features, the presumed useful ones are selected according to our domain knowledge and task needs. 

\subsubsection{Congestion index} The congestion indices of ground and airspace are calculated using flight data and trajectory data, respectively. The ground congestion indices are computed by counting departures and arrivals within 10-minute bins. In a similar way, congestion indices of airport airspace (low-altitude) can be obtained but we use ADS-B data because it contains both commercial airplanes and private jets that makes the results more accurate. To be specific, the flight records whose aircraft-to-ATL distance is less than 200 KM and altitude ranges from 1200 to 10,000 feet above the mean sea level (MSL) are counted. The distance is obtained by Haversine formula given the longitude and latitude of ATL and an airplane:

\begin{equation}
    d=2 r \sin ^{-1}(\sqrt{\sin ^{2}\left(\frac{\Delta\phi}{2}\right)+\cos \left(\phi \right) \sin ^{2}\left(\frac{\Delta\psi}{2}\right)})
    \label{eq:hav}
\end{equation}
\begin{equation}
    \Delta\phi = \phi_{2}-\phi_{1}, \Delta\psi = \psi_{2}-\psi_{1}
    \label{eq:hav2}
\end{equation}
\begin{equation}
    \cos \left(\phi \right)=\cos \left(\phi_{1} \right) \cos \left(\phi_{2} \right)
    \label{eq:hav3}
\end{equation}

\noindent where $d$ is the great-circle distance between two points on a sphere with longitude and latitude ($\phi$, $\psi$) and $r$ is the radius of the Earth. Besides, we split the airspace begins at 18,000 feet above MSL into equal-sized sectors and the number of aircraft in each sector per 10 minutes is defined as the congestion index of en-route airspace.

\begin{figure*}
    \centering
    \includegraphics [scale =.35] {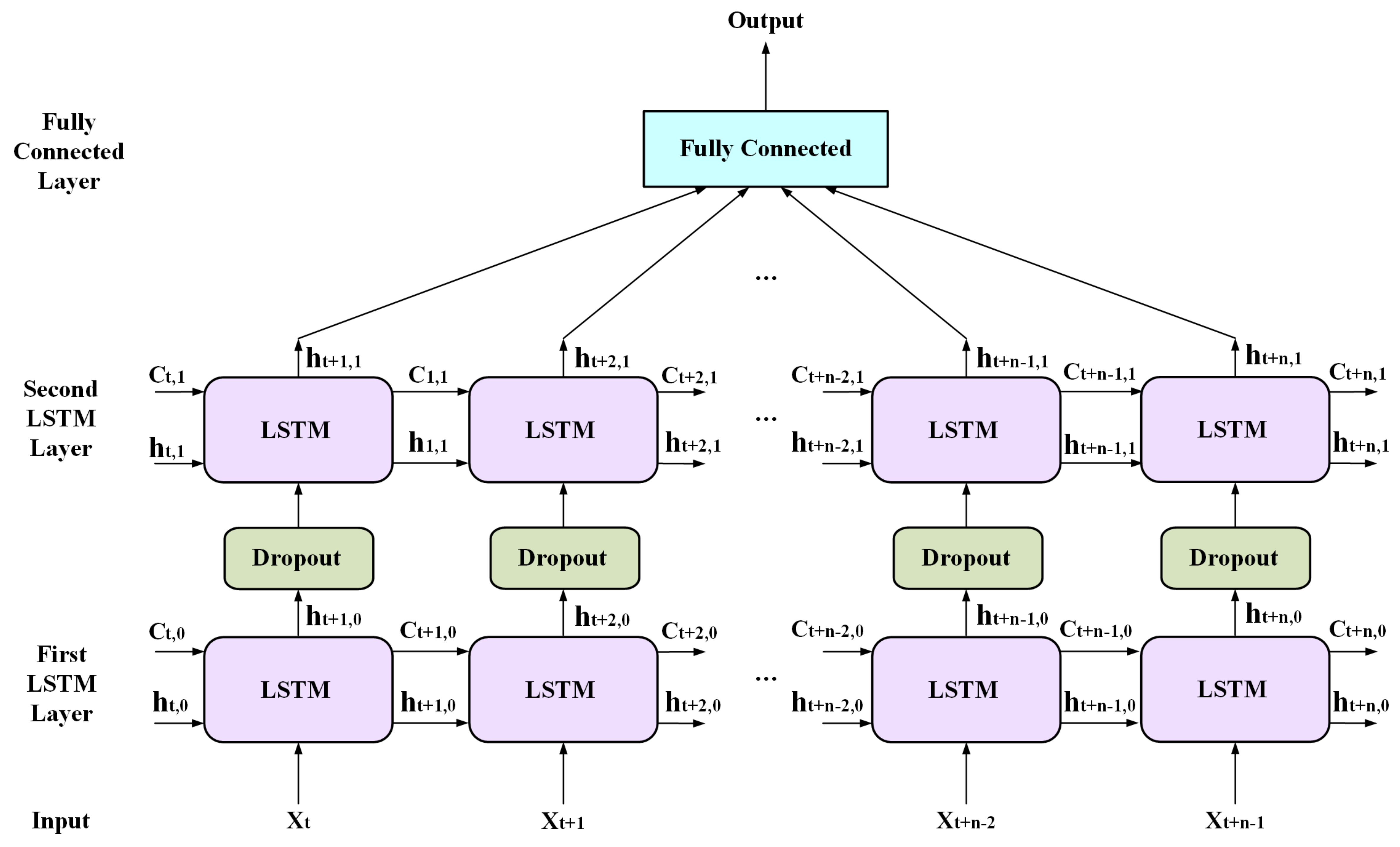}
    \caption{Stacked LSTM architecture with Dropout.}
    \label{fig:lstm}
\end{figure*}

\subsubsection{Encode discrete and categorical features}
To ensure the consistency of the sequence length, we need to summarize the information on succeeding flights because the number of flights are uncertain at every point in time. However, it is inappropriate to take the average of these features because our data set contains some discrete and categorical variables. Considering the interpretation and diversity of these features, a hybrid encoding strategy is adopted in this paper. More narrowly, the discrete and categorical variables are split up into two types -- high-cardinality and low-cardinality based on the threshold of 50. Then, we encode the features using frequency encoding and one-hot encoding, respectively. One-hot encoding enables the representation of the discrete feature to be mapped into Euclidean space, and a certain value of the discrete feature corresponds to a point in the Euclidean space so that it is more reasonable to calculate the distance between them. However, it would cause the curse of dimension if we encode high-cardinality features using one-hot encoding. Hence, we use frequency encoding, which counts and sorts the occurrence of values, to address this issue. 

\subsubsection{Data fusion}
Before we combine various cleansed data sets, we have to convert the timestamps into a unique time zone, UTC. Then, we merge weather data with trajectory data and flight data separately to create two data sets named $D_{t}$ and $D_{f}$, respectively. Recall Section \uppercase\expandafter{\romannumeral2}, here we set $\tau=60$ minutes which means if we assume the timestamp of the last instance in the temporal sequence is $t$, the aviation delay situation at time $t+\tau$ is chosen as the ground truth. However, there may be no aircraft arrives at ATL at the time of $t+\tau$. Thus we take the average delay of aircraft whose arrival times lie between $t+\tau$ and $t+\tau+5$, which could be positive or negative, as the target of our prediction task. It is obvious that airplanes whose estimated arrival times fall within the interval between $t+\tau$ and $t+\tau+5$ are still flying. Therefore, spatio-temporal information of these airplanes such as how close they are to ATL airport, weather situations and their speeds are retrieved from $D_{t}$ and introduced to $D_{f}$ based on aircraft registration identification and time flag. Finally, we fuse ATC data and $D_{f}$ together when taking geolocation as the key value.

\subsection{LSTM Based Architecture}
The success of deep neural networks on a wide range of challenging prediction problems is commonly attributed to the hierarchy and depth \cite{pascanu2013construct, hermans2013training}. Inspired by this, Graves et al. \cite{graves2013speech} demonstrate that RNNs can also benefit from depth in space by stacking multiple recurrent hidden layers on top of each other. Therefore, we adopt stacked LSTM architecture for our aviation delay prediction. Fig. \ref{fig:lstm} shows the structure of our proposed framework. There is a sequence with length $N$, and each output of the second LSTM layer is jointed with a fully connected (FC) layer. We expect the model looks backward historical states and attempts to capture potential temporal characteristics since some important hidden representations may be lost in the last LSTM cell. Besides, FC layer with more neurons has better expressivity for a complex function. In the process of our experiments, we found that the loss converges too quickly 
and tend to 0 which is proven to be the overfitting problem, thus we add a Dropout regularization between two LSTM layers. 



\section{Experiment}
In this section, we present the construction of the time series. Then we evaluate the performance and effectiveness of our LSTM-based architecture compared with other commonly-used machine learning algorithms for aviation delay prediction. 

\subsection{Data Preparation}
\begin{figure}
    \centering
    \includegraphics [width=0.45\textwidth] {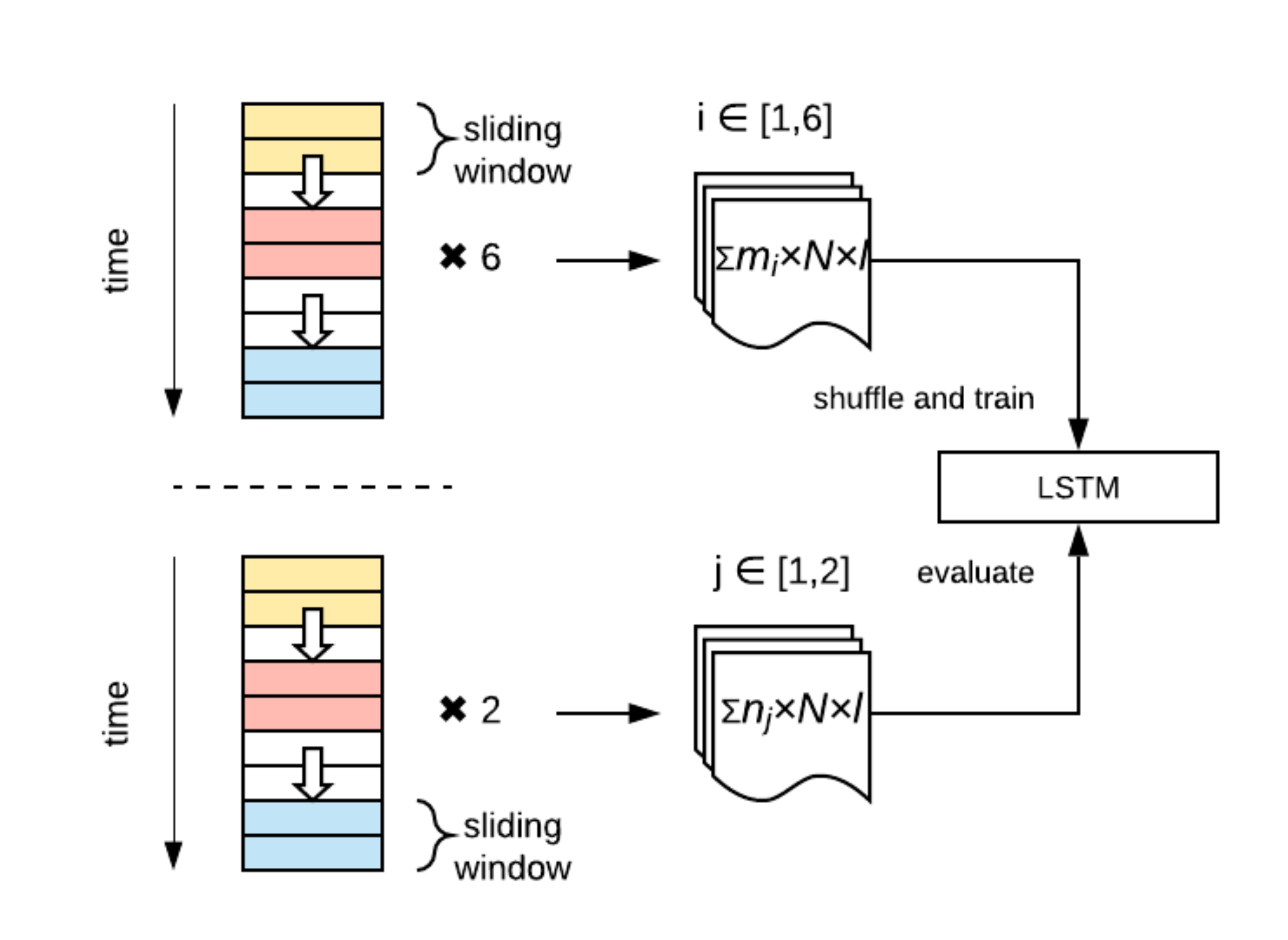}
    \caption{Array slicing for constructing time series. \textit{m} and \textit{n} denote the number of instances of daily data after slicing for training and testing, respectively; \textit{N} represents the length of sliding window over time, and \textit{l} is the number of features.}
    \label{fig:data}
\end{figure}

The inputs for LSTM are 3-dimensional sequential data. Fig. \ref{fig:data} shows the progress of constructing multivariate time series. It should be noted that our data only contains records of the first day of each month from July 2016 to February 2017 in UTC, so we have to split the original data into 8 days by time matching firstly and then slice them separately to create a series of time blocks with a length of \textit{N}. Next, the training set and test set are generated by stacking these arrays in sequences vertically, which keeps the time continuity within a time block. The order of inputs will affect the training results since the former samples will get larger gradients in general, while the neighboring sequences in the time series have the similar distribution. Hence, we shuffle the order in which sequences are fed to the LSTM to guarantee the generalization of the trained model. Note that we do not shuffle the ordering of elements within individual sequences.

\subsection{Setup}
Two types of data sets are used for training and evaluation. The first data set with 66 features contains only flight and weather information in the airport, while the other data set with 203 features contains not only the first data set has but also spatio-temporal information of succeeding flights. We let the sequence length $N$ to be 30, 60, 90, and 120, respectively to generate data sets with different time steps. Mean Square Error (MSE) is chosen as the loss function in the training. 

Our experiment is divided into three steps:

(1) Because most of the researches utilize ensemble methods to predict aviation delay, the first step is to compare the performance of our model and other powerful ensemble algorithms: Random Forest Regression (RF), a bagging method; and Gradient Boosting Regression Tree (GBRT), a boosting method. RF and GBRT are not designed for sequential prediction so we just feed all data into the model instead of slicing data according to $N$ like LSTM. Besides, other commonly-used baseline models such as Linear Regression (LR), Support Vector Regression (SVR), Regression Tree (RT) and Multilayer Perceptron (MLP) are also used for comparison.

(2) The second step is to use two data sets separately to train our stacked LSTM model based on different sequence length, $N$. Therefore, we can analyze the impact of sequence length and en-route spatio-temporal information for our model. 

(3) We apply the model for different airports to verify the validity. The airports include large hub airports such as Los Angeles International Airport (LAX),  O’Hare International Airport (ORD), a mid-size airport, Orlando International Airport (MCO), and a small airport, Daytona Beach International Airport (DAB).

\subsection{Results}
The MSE ($minutes ^2$) is used as the evaluation metric in this paper. Table \ref{tab:t_result} compares performance of different model settings based on $\mathcal{T}$ data. $\mathcal{T}$ represents the data set without information on subsequent flights, on the contrary, $\mathcal{ST}$ represents the entire data set. The best score is highlighted. In a quick glance, it seems clear that our LSTM model outperforms other models. The second best schema is the ensemble method, in which GBRT works best. These conclusions can also be drawn from Table \ref{tab:st_result}. LR perform poorly on both $\mathcal{T}$ and $\mathcal{ST}$ that illustrates there is no apparent linear relation between features and target. Generally speaking, the performance of models with richer spatio-temporal features is better than models that take only temporal correlations into account. Besides, we observe that there is a trend of decline for MSE for test set as the length of the sliding window increases. We think the reason for this phenomenon is that the LSTM learns more hidden patterns of prior knowledge from longer sequences. Although longer sequences may introduce noise to the model during training, the gate mechanism prevents the abuse of long-term dependencies. 

\begin{figure}
    \centering
    \includegraphics [scale =0.45] {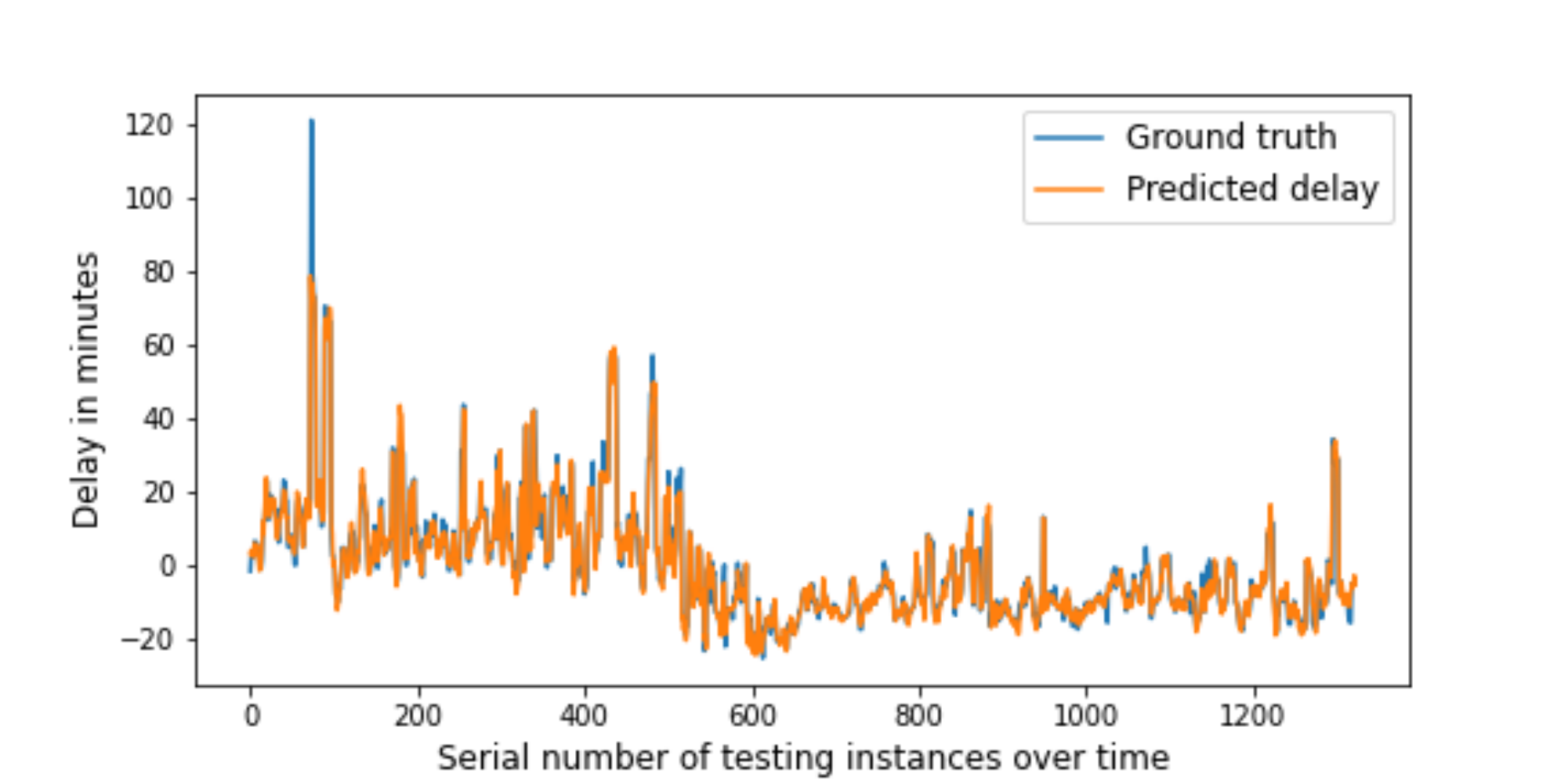}
    \caption{Predicted delay vs. ground truth for $\mathcal{ST}$ test set (LSTM-120)}
    \label{fig:prediction}
\end{figure}

Fig. \ref{fig:prediction} depicts the gap between the predicted value and the ground truth over time in a more intuitive way. It can be clearly seen that there is a very slight difference in amplitude, and their trends are consistent basically which also verifies that our proposed model possesses the ability to predict aviation delay accurately.

Table \ref{tab:airport} shows the MSE values of test sets for different airports. The distributions of data for different airports vary dramatically no matter from spatio-temporal perspective (e.g. weather varies in different cities) or from flight's perspective (e.g. an airline has different flight arrangements for different airports). Therefore,  we use the data at various airports to re-generate time series with spatio-temporal features and re-trained the LSTM-120 model. It demonstrates the effectiveness of our proposed LSTM model and the corresponding feature engineering for large hub airports. However, this model does not make sense for mid- and small-size airports due to the relatively few flights, which are hard to generate representative sequences.   

\begin{table}
    \centering
    \caption{Comparison of prediction accuracy using $\mathcal{T}$ data and different sequence length}
    \setlength{\tabcolsep}{4mm}
    \begin{tabular}{llc}
    \toprule 
    \textbf{Method} & \textbf{Model} & \textbf{Test MSE} \\
    \toprule 
    Linear & LR & 218.9779 \\
    \midrule 
    \multirow{2}{6em}{Non-linear} & RT & 111.2693 \\
    & SVR & 150.2401 \\
    \midrule
    \multirow{2}{6em}{Ensemble} & RF & 84.3054 \\
    & GBRT & 68.9447 \\
    \midrule
    \multirow{5}{6em}{Neural Network} & MLP & 119.0170 \\
    & LSTM-30 & 64.1153 \\
    & LSTM-60 & 62.3336 \\
    & LSTM-90 & 60.3336 \\
    & \textbf{LSTM-120} & \textbf{52.7232} \\
    \bottomrule
    \end{tabular}
    \label{tab:t_result}
\end{table}

\begin{table}
    \centering
    \caption{Comparison of prediction accuracy using $\mathcal{ST}$ data and different sequence length}
    \setlength{\tabcolsep}{4mm}
    \begin{tabular}{llc}
    \toprule 
    \textbf{Method} & \textbf{Model} & \textbf{Test MSE} \\
    \toprule 
    Linear & LR & 202.3007 \\
    \midrule 
    \multirow{2}{6em}{Non-linear} & RT & 77.2012 \\
    & SVR & 57.7634 \\
    \midrule
    \multirow{2}{6em}{Ensemble} & RF & 52.0770 \\
    & GBRT & 43.8420 \\
    \midrule
    \multirow{5}{6em}{Neural Network} & MLP & 56.8772 \\
    & LSTM-30 & 43.9821 \\
    & LSTM-60 & 39.5061 \\
    & LSTM-90 & 35.7900 \\
    & \textbf{LSTM-120} & \textbf{25.6320} \\
    \bottomrule
    \end{tabular}
    \label{tab:st_result}
\end{table}

\begin{table}
    \centering
    \caption{Comparison of prediction accuracy of LSTM-120 using $\mathcal{ST}$ data at different airports}
    \setlength{\tabcolsep}{4mm}
    \begin{tabular}{cc}
    \toprule
    \textbf{Airport} & \textbf{Test MSE} \\
    \midrule
    ATL & 25.6320 \\
    LAX & 27.3342 \\
    ORD & 46.7891 \\
    MCO & 73.5672 \\
    DAB & 110.6825\\
    \bottomrule
    \end{tabular}
    \label{tab:airport}
\end{table}

\section{Conclusion and Future Work}
In this work, we present a novel aviation delay prediction framework based on stacked LSTM for commercial flights. We provide a complete data processing pipeline and generate a data set with richer spatio-temporal features while regarding delay propagation and uncertainty in the flights. Our experiments verify that our framework can predict a commercial flight arrival delay in the US within an acceptable error.  

There is some significant work needed to do in the future. Firstly, we should collect more data to improve the robustness of our proposed model. Moreover, due to the difference in the number and frequency of flights in airports, in particular small airports which only hold a small volume of data that is not enough to train a high-performance but data-greedy deep learning model, this is a strong motivation to implement transfer learning to improve model performance for such airports with the help of prior knowledge. 

\section*{Acknowledgment}

This research was supported by the Center for Advanced Transportation Mobility (CATM), USDOT Grant \#69A3551747125.

\bibliographystyle{./bibliography/IEEEtran}
\bibliography{./bibliography/IEEEabrv,./bibliography/IEEEexample}

\end{document}